\documentclass[preprint,5p,twocolumn]{elsarticle}

\usepackage{hyperref}
\usepackage{graphicx}
\usepackage{amsmath}
\usepackage{amssymb}
\usepackage{multirow}
\usepackage{diagbox}
\usepackage{caption} 
\usepackage{subcaption}
\usepackage{float}
\usepackage{stackengine}
\usepackage{dashrule}

\usepackage{amsmath,amssymb,enumitem,mathtools}

\newcommand{\NA}{--}

\journal{Computers in Biology and Medicine}

\begin{document}

\begin{frontmatter}

\title{Time-based Self-supervised Learning for Wireless Capsule Endoscopy}


\author[aff:ub]{Guillem Pascual\corref{corr}}
\ead{guillem.pascual@ub.edu}
\cortext[corr]{Corresponding author}

\author[aff:ub]{Pablo Laiz}
\author[aff:ub]{Albert García}
\author[aff:hagen]{Hagen Wenzek}
\author[aff:ub]{Jordi Vitrià}

\author[aff:ub]{Santi Seguí}

\address[aff:ub]{Department of Mathematics and Computer Science, Universitat de Barcelona, Barcelona, Spain}
\address[aff:hagen]{CorporateHealth International ApS, Denmark}

\begin{abstract}
State-of-the-art machine learning models, and especially deep learning ones, are significantly data-hungry; they require vast amounts of manually labeled samples to function correctly. However, in most medical imaging fields, obtaining said data can be challenging. Not only the volume of data is a problem, but also the imbalances within its classes; it is common to have many more images of healthy patients than of those with pathology. Computer-aided diagnostic systems suffer from these issues, usually over-designing their models to perform accurately. This work proposes using self-supervised learning for wireless endoscopy videos by introducing a custom-tailored method that does not initially need labels or appropriate balance. We prove that using the inferred inherent structure learned by our method, extracted from the temporal axis, improves the detection rate on several domain-specific applications even under severe imbalance.
\end{abstract}

\begin{keyword}
 capsule endoscopy \sep deep learning \sep self-supervised learning \sep semi-supervised learning
\end{keyword}

\end{frontmatter}


\section{Introduction}

\begin{figure}
    \centering
    \includegraphics[width=0.49\textwidth]{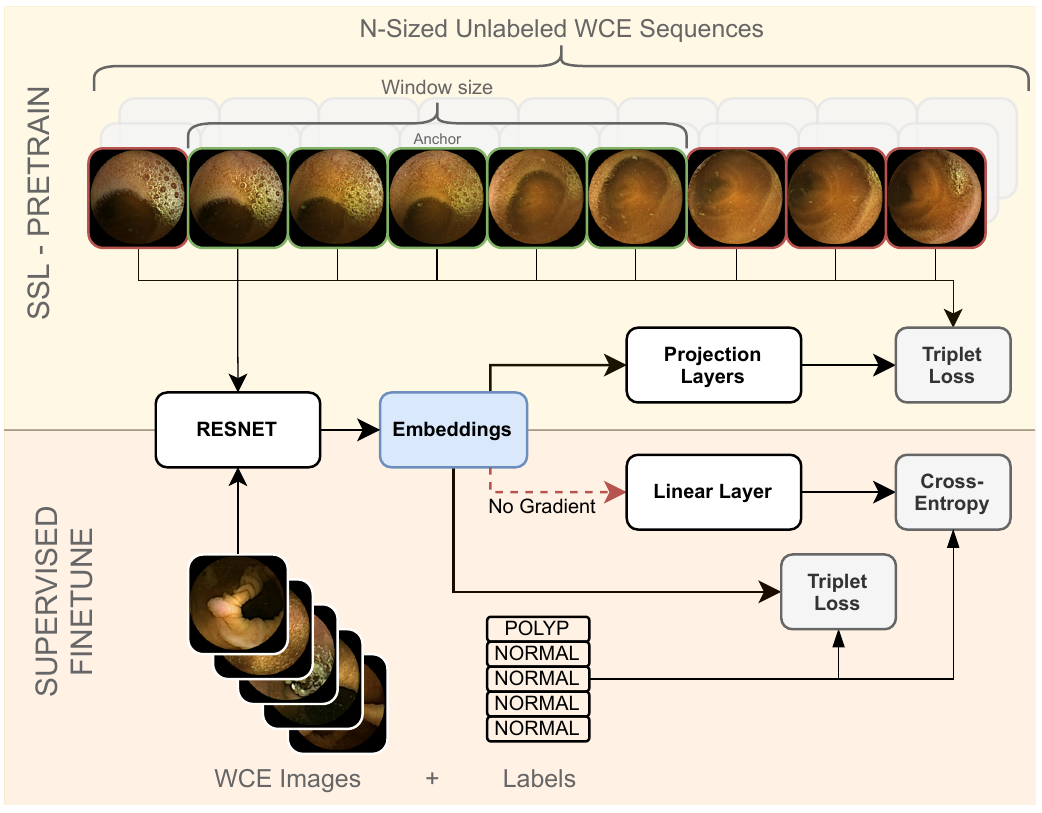}
    \caption{Overview of the proposed method, including the pretain phase, in the upper half, and the final finetune phase in the lower half.}
    \label{fig:splash}
\end{figure}%

Obtaining gastrointestinal (GI) images has traditionally been an intrusive intervention until the advent of Wireless Capsule Endoscopy (WCE) technology \cite{Iddan2000b}. WCE imaging eases the process of securing a continuous stream of images, but at the same time, it introduces its own set of problems.

The videos recorded by the capsule, although usually with a low frame rate, can have a duration of up to 12 hours \cite{Vasilakakis2019}. Unlike traditional methods, it is not a targeted exploration but rather a complete recording as the capsule travels through the entire system. A physician must go over the full length of the video, possibly at multi-image speeds, while looking for any abnormality. Not only have they to invest considerable more time, but the fatigue and repetitiveness of the task could affect their ability to detect such abnormalities.

Providing a reliable and accurate computer-aided diagnosis (CADx) system capable of selecting the most promising frames would ease the pressure for those professionals, cutting down the time spent on the task while obtaining comparable---if not better---results.

Also of great importance, especially when designing automated systems that rely on images obtained from patients, is to examine the properties of the data. In day-to-day examinations, not all patients have an associated pathology, and the data used in research to train CADx models directly reflects it. In polyp detection, for example, the majority of videos have no polyp present in a video at all. One must also consider that, even in the case that there might be polyps, they would appear only in a small fraction of the frames \cite{Laiz2020a}. A polyp might appear in several subsequent frames, perhaps slightly displaced or rotated, but the overall number would be negligible when considering the whole duration of the video.

The problem statement becomes clear when one combines the aforementioned problem with the difficulty of obtaining said datasets. Data is fairly scarce compared to other problems studied in deep learning, and the classes, such as polyp, or no polyp, suffer significant imbalances. Not to mention that supervised algorithms, which dominate the field, require that all those videos are accurately labeled to function.

As laid out in the next section, existing methods have mostly focused on traditional classification tasks. They have to work with low amounts of highly imbalanced data, relying on techniques like data augmentation and regularization to cope with overfitting and under-performing models.

Instead, this work proposes a novel application of self-supervised learning (SSL) to obtain a better representation of the data, enabling future models to perform better in their classification tasks. Self-supervision, although mainly relying on unlabeled data,  has  been canonically considered a variant of supervised learning \cite{Liu2021a}. The network learns from supervisory signals obtained from the data itself, often leveraging the underlying structure in the data. While it requires some pseudo-labeling on the images, it can be often be extracted from the data itself.

In SSL, instead of directly training a model with a set objective in mind, the process is divided into two steps. SSL is done during during an initial phase named {\em pretrain}, where a deep neural network is trained to learn a better representation, or embedding, of the data. It encodes the most essential information into a smaller vector by using the data without their final labels, learning its inherent structure. This information is learned accordingly to the data's nature, the model's architecture, and the task used for SSL. Then, during a second pass, the {\em finetune} process, the embedding is used in conjunction with the labels to perform supervised classification. 

With the present work, summarized in Figure \ref{fig:splash}, we aim to use self-supervision to provide more accurate models for domain-specific tasks derived from WCE images. In particular, given unlabeled WCE videos, we exploit their temporal nature to perform SSL and then train several supervised models.

The paper is organized as follows. First, we give an overview of the related work in the field followed by a description of our methodology, presenting the self-supervised training, supervised training, and system architecture. Further, we explain the experimental setup and results, and finally present the main conclusions and give directions for future work.
\section{Related work}

Given the nature of our work, which covers standard classification methods, medical imaging, and SSL, the section is structured in three subsections. First, we review traditional WCE classification models, focusing on those attempting to create models via traditional methods or deep learning. Then, stepping into supervised training, recent works in the field and applications in WCE are shown. Finally, SSL publications are explored.

In regards to WCE imagery, one of the first notable works \cite{Fu2014} used superpixels in conjunction with a support vector machine to detect bleeding. In fact, bleeding detection is one of the most explored domains in WCE imaging and one of the most successful \cite{Pogorelov2019,Lv2011,Yuan2015}. Deep learning has also been recently used in other tasks, such as motility events analysis \cite{Segui2016}, polyp detection \cite{Aoki2019,Yuan2015}, and ulcer detection \cite{V2020}.

As has been previously stated, one problem all of the above methods suffer is the lack of labeled data. Some domain-specific tasks, like polyp detection, also have highly imbalanced data. This is formalized and analyzed in \cite{Yuan2020,A2019} works, where the difficulties derived from imbalance, low inter-class variance, and high inter-class variance are inspected in detail. Techniques like dropout, L1 or L2 regularization, and sampling mechanisms have been applied to attempt to soften the problem \cite{Kim2021}.

Other approaches to tackle such issues are self-supervision methods, of which a wide range of options are available. For instance, a popular architecture choice was autoencoders \cite{Rumelhart2013,Kingma2014,Hinton2011}, whose dimensionality-reducing capabilities were believed to be useful for SSL. However, it has been demonstrated that they fail to capture rich information \cite{Bengio2009}, focusing only on compressing data. Thus, their capacity to adapt to any future generic task is hindered at best.

In contrast to the former generative method, where the network learns from a single image, contrastive learning trains on multiple examples or instances of the same image to learn the inherent information \cite{Falcon2020}. One such way to introduce multiple samples of a single image has been by reordering subsections \cite{Misra2019}. This type of SSL encourages the network to learn invariant representations, unlike their generative counterparts. Similarly, when the time dimension is available, reordering can be done based on fragments of the input, as done with audio streams \cite{Oord2018a}.

Additional techniques, like rotation, color jittering, blurring, and cropping, can be applied as shown in \cite{Chen2020b, Chen2020c}. The authors propose SimCLR, an architecture based on ResNet \cite{He2016a} that can be trained with multiple contrastive approaches and a new contrastive loss. They provide a simple framework to perform SSL and benchmark the different methods.

More specifically and related to our application, SSL from videos has been done by predicting the order of a sequence \cite{Misra2016,Xu2019,Lee}, object tracking \cite{Pathak2016,Wang,Wang2019}, and contrastive losses \cite{Tschannen2019,Sermanet2018}. In particular, our method resembles the single-view approach of Time-Contrastive Networks \cite{Sermanet2018}, however, their work diverges from ours because they do not focus on the embeddings' richness nor task-generalization. Given the nature of their action imitation task, they limit their triplets to be in a single sequence and do not explore the embedding quality, whereas our work aims to learn generalized and rich embeddings from hours-long videos, exploring inter-sequence and inter-video triplets.

Some efforts have been made in regards to SSL and semi-supervised training in medical imagery \cite{Perez-Garcia2021,Cheplygina2019,Azizi2021}, proving that it can improve results in tasks such as pneumonia detection and multi-organ segmentation \cite{Navarro2021}. It has also been applied to WCE related tasks \cite{Guo2020a,Vats2021}, although there has not been any work that, to the best of our knowledge, has leveraged the temporal aspect of WCE videos.

\section{Method}
An overview of our proposed self-supervised approach is illustrated in Fig. \ref{fig:splash}. Similar to most methods relying on self-supervised training, our approach is divided into two distinct stages: (a) pretraining a self-supervised network using unlabeled data to obtain rich representations, and (b) finetuning the model using labeled data for a specific task. This section follows the same pattern, explaining both phases first, and finishes by explaining the architecture used.

\subsection{Self-supervised pretraining}

During the first stage of the process, we aim to extract useful generic information from the unlabeled images, which then can be transferred to deal with many specific tasks by finetuning the model with limited labeled data. In other words, it creates a reduced representation (embedding) of the original image that contains its most important information. 

Extracting an embedding can be understood as a process $f(x)$, where a neural network transforms a sample $x$ from the dataset to its compressed and rich representation.

Out of all the possible ways to obtain said embedding, we have chosen to exploit the temporal nature of WCE videos. Our method works by taking sequences of $N$ contiguous frames and creating a relationship between them. Namely, given two frames $i, j$ in the sequence, their relationship is established as the distance $d(i, j)$ between them, counted by the number of frames that separates them.

Unlike the work in \cite{Sermanet2018}, where all samples come from a single sequence, our method must generalize to multiple videos and sequences. Per-frame pseudo-labels are introduced to encode their video identifier along with their position. Given an image $i$, its pseudo-label is a combination of its video identifier $\gamma(i)$, which can be a simple numbered sequence, and the position inside the video $\delta(i)$, as seen in Equation \ref{eqn:label}.

\begin{equation}
\label{eqn:label}
\bar{y}(i) = M\gamma(i) + \delta(i)
\end{equation}

Where $M$ must be a large enough number so that $ \forall i, M > \delta(i)$. For our particular experiments and datasets, we have chosen $M = 10^6$.

Next, we impose a similarity measure between frames on the sequence so that contrastive learning can be done by finding the inherent relationship between similar and dissimilar images. For that purpose, two images will be consider similar if they are close enough, formalized as $d(i, j) = \lvert\bar{y}(i)-\bar{y}(j)\rvert \leq w$, where $w\leq N$ is a constant chosen beforehand. The pair $(i, j)$ is considered similar (positive) in such cases, and negative otherwise.

In other words, taking a reference image (anchor) in a sequence, all other images within a window of size $2w$ ($w$ images per side) are considered similar. In general, given an $N$-sequence, all images have between $min(N, 2w)$ and $w$ positive samples. Images around the edges of the sequences lose up to half the positives, tending towards the latter, while those on the center have the whole spectrum.

The pseudo-labels guarantee that $(i, j)$ negative pairs are consistent with images coming from different videos, as $\gamma(i)\neq\gamma(j)$, thus $d(i,j) \approx \lvert M\gamma(i) - M\gamma(j)\rvert\geq M > w$. Additionally, for two frames $i,j$ extracted from the same video, the formula reduces to the distance in frames between them, $d(i, j) = \lvert\bar{y}(i) - \bar{y}(j)\rvert = \lvert\delta(i) - \delta(j)\rvert$.

Given the above approach to create a similarity measure, the Triplet Loss (TL) \cite{Schultz}, a contrastive loss, is introduced to learn the embeddings. TL works by using triplets of samples, where two of the triplet's elements, the anchor $a$ and the positive $p$, pertain to the same class. The remaining element, the negative $n$, is of a different class than $a$. That is, given the embedding of an anchor $f(a)$, a triplet $(f(a), f(p), f(n))$ is formed so that $y(a) = y(p) \neq y(n)$, where $y(\cdot)$ is the class of a sample. 

Using Equation \ref{eqn:tl}, TL forces $f(p)$ to be close to $f(a)$ while moving away $f(n)$. It eases the problem by introducing a soft margin $\alpha$ between the positive and negative pairs.

\begin{equation}
\label{eqn:tl}
TL = max(||f(a) - f(p)||^2 - ||f(a), f(n)||^2+\alpha, 0)
\end{equation}

Translated to our domain, a triplet is formed by two similar images and a dissimilar image, so that $d(a, p) \leq w$ and $d(a, n) > w$. As shown, TL is directly applicable to WCE videos when used in conjunction with the pseudo-labels, forcing close images in a sequence to have similar representations in the embedding space.

It must be noted that this method is bound to have incorrect pairs, as different videos or sequences could contain similar images, regardless of their distance. Also, WCE videos tend to have periods where the capsule moves at a slow rate, producing many similar images in a relatively long interval, or the contrary, moves fast and captures rapidly changing sequences. We estimate those cases to be negligible compared to our dataset's size, being effectively treated as noise during the process.

\subsection{Supervised learning}

During the second phase of our method, the same model is reused to learn a domain-specific task with limited amounts of data. For instance, the rich representations could be used to model motility events, to classify several conditions like bleeding or inflammation, to evaluate keyframes, or to detect polyps, to name a few.

For that purpose, the process starts with the SSL model's parameters, obtaining embeddings produced by the new dataset and feeding them into a classifier. That classifier needs to access the ground truth labels, as it uses a softmax cross-entropy loss to model the problem.

Following SimCLR findings \cite{Chen2020b}, we have confirmed that fixing the weights obtained during SSL is counterproductive. However, unlike SimCLR, which assumes balanced problems, we use the approach proposed by Laiz \textit{et al.} \cite{Laiz2020a}, where the TL is used to modify the embeddings. As such, the gradient coming from the linear classifier is removed so that it cannot negatively impact the embeddings due to the imbalance. Instead, a TL is imposed on them to facilitate the network to finetune the dataset representations.

However, unlike in the previous step, the TL no longer uses the pseudo-labels created through our method. Triplets are formed by considering the real labels of the images, which are domain-specific and help finetune the embeddings to the particular task. To further reference it and avoid confusion, the term $TL_{sup}$ will be used.

The $TL_{sup}$ is trained in batch all mode, which considers all triplets regardless of their difficulty. No special sampling algorithm is introduced; the only restriction we impose is for a batch to have a proportional representation from all classes. Other than that, data is randomly sampled.

The final loss obtained in this model is the linear combination of both the cross-entropy loss and the triplet loss, as shown in Equation \ref{eqn:supervised-loss}.

\begin{equation}
    L_{sup} = TL_{sup} + L_{crossentropy}
    \label{eqn:supervised-loss}
\end{equation}

\subsection{Architecture}

\begin{figure}
    \centering

    \begin{subfigure}{.25\textwidth}
        \centering
        \includegraphics[width=.9\linewidth]{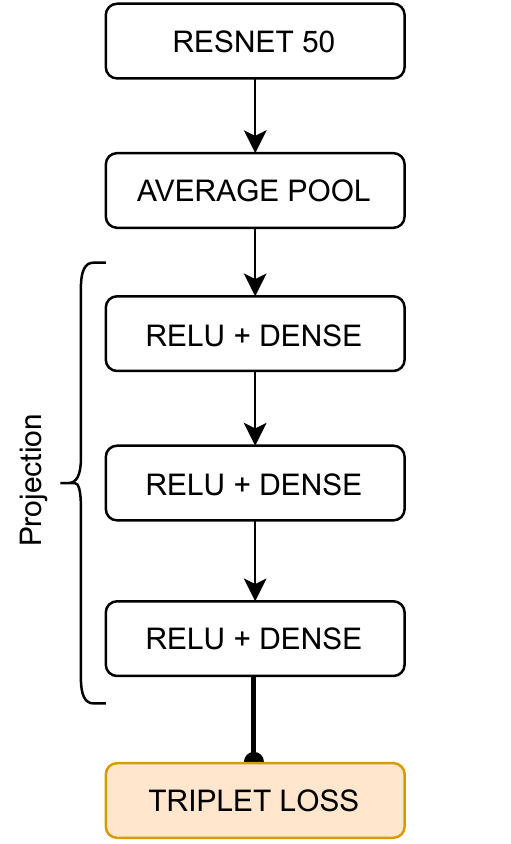}
        \caption{Pretrain}
        \label{fig:pretrain}
    \end{subfigure}%
    \begin{subfigure}{.25\textwidth}
        \centering
        \includegraphics[width=.9\linewidth]{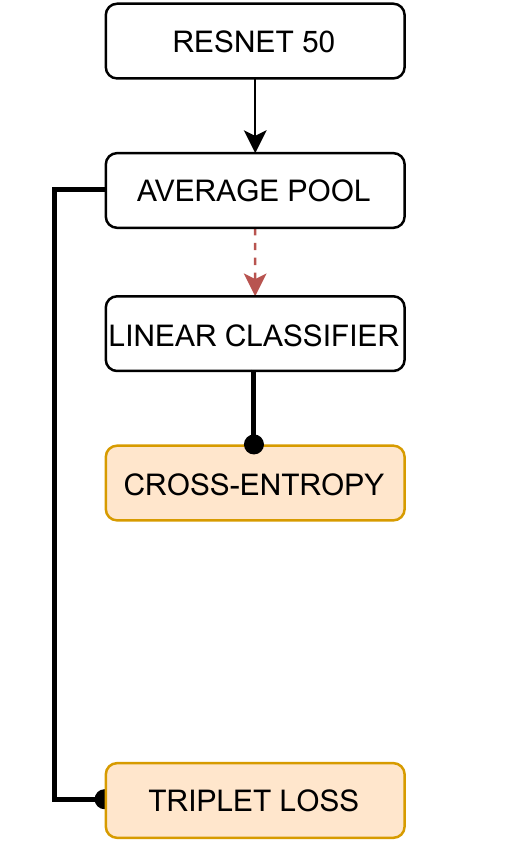}
        \caption{Finetune}
        \label{fig:finetune}
    \end{subfigure}
    
    \caption{Detailed network architecture. The parameters obtained during pretain for ResNet are used in the finetune phase, while the projection layers are removed. Here, the dashed red line denotes that gradient is stopped.}
    \label{fig:test}
\end{figure}

The backbone of our architecture consists of a ResNet-50 \cite{He2016a}, as can be seen in Figure \ref{fig:pretrain}. Most works that extract or require embeddings use the output of the ResNet model directly as their representations, but following the work in SimCLR, we decided to explore the possibility of including several projection layers.

Each projection layer consist of a ReLU activation followed by a dense layer. We restrict all the projection layers to have the same dimensionality, which must be lower than the 2048 given by ResNet. While our pretrain phase benefits from the reduced complexity after the projection, the final finetuning network utilizes the whole 2048-sized embedding to allow for better detection rates. These layers, along with their configuration, hyperparameters, and performance, are studied below. Ultimately, they are found to be beneficial for domain-specific tasks.

Once the pretrain is done, at the beginning of the finetune phase all learned parameters are kept except for the projection layers, which are removed from the model, as can be observed in Figure \ref{fig:finetune}. Classification is done through a linear layer (a dense layer without any activation) and a cross-entropy loss. As denoted in red and a dashed line in Figure \ref{fig:finetune}, we eliminate the gradient coming from the linear classifier to stop it from modifying the embedding. Only the TL loss is able to tune the representations.

It must be remarked that the TL losses used in both phases of the architecture are different. As pointed out, the first phase uses the pseudo-labels deducted from videos, while the second uses the ground truth labels.

\begin{table*}[t]
    \caption{Hyperparameters tested during the self-supervised training, combining different Sequence Sizes ($N$) and Window Sizes ($w$). Resampling indicates that, in a single batch, all sequences come from the same video. Note that resampling only makes sense if $N$ is smaller and multiple of the batch size.}
    
    \centering
    
    \begin{tabular}{ccccc}
    \hline
    \textbf{Sequence Size} & \textbf{Sequences per Batch} & \textbf{Window Size} & \textbf{Resample} & \textbf{AUC (\%)}           \\ \hline
    9                      & 8                            & 3                    & No                & $93.51\pm1.35$         \\
    9                      & 8                            & 3                    & Yes               &   $93.23\pm1.78$        \\
    9                      & 8                            & 6                    & No                & $93.49\pm1.31$         \\
    9                      & 8                            & 6                    & Yes               &   $93.81\pm2.12$         \\
    18                     & 4                            & 3                    & No                & $93.68\pm1.97$         \\
    18                     & 4                            & 6                    & No                & $93.47\pm1.11$         \\
    18                     & 4                            & 6                    & Yes               & $92.91\pm2.70$         \\
    18                     & 4                            & 9                    & No                & $93.42\pm1.62$         \\
    18                     & 4                            & 9                    & Yes               & $93.62\pm1.63$         \\
    72                     & 1                            & 6                    & \NA                & $94.12\pm1.35$         \\
    \textbf{72}            & \textbf{1}                   & \textbf{9}           & \textbf{\NA}       & $\mathbf{94.60\pm1.15}$ \\
    72                     & 1                            & 18                   & \NA                & $94.14\pm2.12$         \\
    72                     & 1                            & 32                   & \NA                & $94.53\pm0.96$         \\ \hline
    \end{tabular}
    
    \label{table:pre-hyper}

\end{table*}

\section{Discussion and results}

This section begins by laying out the datasets used during both steps of the method. Further, it explains the implementation details, such as preprocessing steps and train strategies. A subsection is devoted explicitly to the SSL hyperparameters, justifying and proving the choices made. Finally, individual results are shown for each dataset, discussing the results qualitatively and quantitatively.

\subsection{Datasets}
\subsubsection*{Generic WCE videos}
This dataset consists of a total 49 unlabeled WCE videos, each from different patients, obtained with Medtronic PillCam SB3 and PillCam Colon 2. From those videos, only the small intestine and colon segments are used, selecting a total of 1,185,033 frames with a resolution of 256 by 256 pixels. 

\subsubsection*{Polyp WCE}
The dataset consists of 248,136 frames sampled from Medtronic PillCam SB3 and PillCam Colon 2 videos. Notably, they are not the same videos as the subsection above. Of those frames, 2,080 contain polyps, while 246,056 do not. As can be seen, this dataset suffers from the exact problems this publication aims to tackle: only 0.85\% of all images contain polyps. It is a highly imbalanced problem with an objectively low amount of samples compared to traditional deep learning settings.

\subsubsection*{CAD-CAP WCE}
This public dataset was compiled during the Gastrointestinal Image ANAlysis (GIANA) challenge \cite{Dray2018}. It consists of three balanced classes: normal, inflammatory, and vascular lesion, each with approximately 600 images for a total of 1,800 images. 

\subsection{Implementation Details}

We performed all the experiments on one NVIDIA Titan Xp GPU, implementing the entire architecture in TensorFlow 2.4. The backbone network, a ResNet-50, was initialized using the Imagenet trained model, while the projection layers were randomly initialized. 

\subsubsection*{Preprocessing} 
All data, including the used in pretrain and finetune, was processed using standard data augmentation (DA) techniques, such as color jittering, grayscale conversion, and random rotations and flips.

We also introduced a mask with a radius of 128 pixels to eliminate any artifacts present at the borders of the images, making sure that no specific noise or patterns could identify either a dataset or a particular video.

For our finetune step, as is customary in the field due to the low number of images, the use of DA is mandatory to avoid overfitting. We found that not introducing this same augmentation on the pretrain step negatively affected our final classification results. Thus, all sections below assume the use of DA techniques for training.

\subsubsection*{Self-supervised learning}
The unlabeled Generic WCE videos were used as training data during this stage. The network was optimized using stochastic gradient descent, without momentum, for a total of 21,000 batches with 72 images each (about 2 hours and 30 minutes on our GPU). In our best-performing configuration, the network processes 21,000 sequences. The learning rate was fixed to 0.1, and was divided by 5 every 4300 iterations. Throughout the process, we used an L2 weight decay of 0.0001. We experimented with multiple values, reaching the same conclusion as SimCLR \cite{Chen2020b}, whereas any low value helps regularize the embedding pre-projection. Finally, we used a batch all strategy for triplet loss, with unnormalized embeddings and a margin of 0.2.

To select the best set of hyperparameters that our SSL method requires, a procedure has been devised. For a particular set of hyperparameters, the network is trained, finetuned over the polyp dataset, and finally evaluated using Area Under the Curve (AUC) computed from Receiver Operating Characteristics (ROC). Unlike a proper evaluation for the dataset itself, which is done through complete videos, as our aim is to validate the hyperparameters only, the scores are extracted through a five-fold cross-validation over randomly selected samples from the videos.


\begin{table*}[t]
    \caption{Study of the effect of adding several projection layers with a varying number of parameters. Each projection layer consists of a ReLU activation followed by a dense layer. All dense layers have the same amount of parameters (dimensionality).}
    
    \centering
    
    \begin{tabular}{ccc}
    \hline
    \textbf{Projection Layers} & \textbf{Projection Dimensionality} & \textbf{AUC (\%)}            \\ \hline
    0                 & \NA                       & $92.97\pm1.19$ \\
    1                 & 128                       & $93.02\pm1.39$ \\
    2                 & 128                       & $94.09\pm1.28$ \\
    \textbf{3}        & \textbf{128}              & $\mathbf{94.60\pm1.15}$ \\
    3                 & 256                       & $93.56\pm1.53$ \\
    6                 & 128                       & $93.85\pm1.80$ \\ \hline
    \end{tabular}

    \label{table:proj}
    
\end{table*}

\subsubsection*{Supervised learning} 
The entire pretrained network was finetuned with a linear classifier on top of the learned representation. All datasets were equally trained with a learning rate of 0.01, decaying it by 10 every 1,500 iterations for a total of 4,500 steps.

\subsection{SSL Hyperparameters}

We first performed experiments to choose the sequences' length $N$, window size $w$, and whether multiple videos should be used in a single batch or not. Due to our available GPU memory, we could fit at most 72 images in a single batch, which set an upper bound to $N$. We designed several models, see Table \ref{table:pre-hyper}, to select the best performing combination. Although the results show no statistically significant difference among some, it can be observed that sequences of 72 images, where all images come from the same video, tend to give better results.

Most images will be relatively similar and close when using a continuous stream of 72 images from a single video. Therefore, triplets formed for TL will consist of hard negatives, namely from samples that are difficult to distinguish. Oppositely, mixing several videos in a single batch will produce negatives that are too easy to distinguish from their anchors.

We believe this added difficulty, albeit making the training process slower, helps the network extract more meaningful information of the images. Thus, richer embeddings are produced, which can then perform better in later downstream tasks. For future experiments, $N$ was fixed to 72, obtained continuously from a single video, and $w$ to 9 images.

Following, we pinpointed the benefits of adding projection layers. We verified, as can be observed in Table \ref{table:proj}, whether adding these additional parameters during the pretraining phase yielded better results during polyp detection. It is of particular importance to remark that any projection layer added is then removed during the second phase, thus the same number of parameters is kept regardless of the choices made here. 

Particularly, the optimal combination for our particular task seemed to be at 3 layers, each of 128 parameters, which yields a substantial improvement compared to using none and outperforms more complex solutions.

\begin{table*}[t]

    \caption{Performance comparison of several methods with the same parameter count. Imagenet refers to a ResNet-50 pretrained on the imagenet dataset and then finetuned with a cross-entropy loss over our dataset. SimCLR has been trained with NT-Xent as per Chen \textit{et al.} \cite{Chen2020b}. $\text{TL}_\text{BA}$ is equivalent to Imagenet but trained with an additional triplet loss. Ours is the self-supervised network.}
    
    \centering
    
    \begin{tabular}{cccccccc}
    \hline
    \textbf{}             & \textbf{AUC}   & \multicolumn{3}{c}{\textbf{Sensitivity \%}}                              \\
    \textbf{Model}        & \textbf{(\%)}  & \textbf{Spec. at 95\%} & \textbf{Spec. at 90\%} & \textbf{Spec. at 80\%} \\ \hline

    Imagenet                & $82.85\pm5.72$ & $37.75\pm9.12$         & $51.49\pm11.09$        & $66.71\pm12.15$        \\
    SimCLR \cite{Chen2020b}                  & $92.76\pm1.62$ & $68.13\pm6.37$         & $76.92\pm5.40$        & $87.91\pm3.94$        \\
    $\text{TL}_\text{BA}$ \cite{Laiz2020a} & $92.94\pm1.87$ & $76.68\pm4.93$         & $82.86\pm4.78$         & $88.53\pm3.76$         \\
    \textbf{Ours}                    & $\mathbf{95.00\pm2.09}$ & $\mathbf{80.16\pm6.97}$        & $\mathbf{86.31\pm6.20}$         & $\mathbf{92.09\pm4.63}$        
    \end{tabular}

    \label{table:results}

\end{table*}

\subsection{Results}

In this subsection, first the quality of the embeddings learned during the self-supervised learning is evaluated. Then, we explore the results obtained with two downstream specific tasks.

\subsubsection*{SSL embeddings}

As stated, our SSL process aims to learn rich embeddings. To such end, we use the temporal sequences extracted from WCE videos to make the network learn when two images are close or not in the video. It is expected that two embeddings of consecutive images are similar.

\begin{figure*}
    \centering
    \stackinset{l}{57pt}{t}{0in}%
    {\rotatebox{90}{\hdashrule{62pt}{0.5pt}{0.3mm}}}{%
        \includegraphics[width=0.99\textwidth]{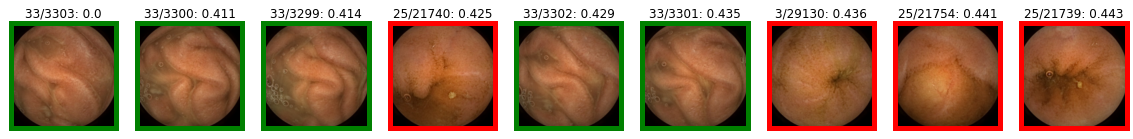}
    }%
    \hspace{-\linewidth}\rule{\linewidth}{0.5pt}
    \stackinset{l}{57pt}{t}{0in}%
    {\rotatebox{90}{\hdashrule{62pt}{0.5pt}{0.3mm}}}{%
        \includegraphics[width=0.99\textwidth]{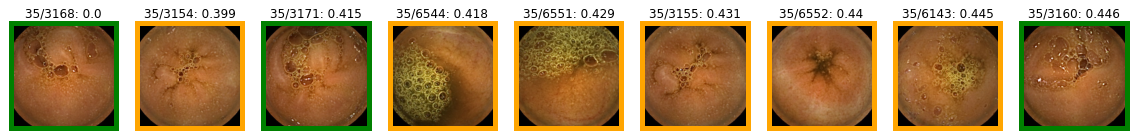}
    }%
    \hspace{-\linewidth}\rule{\linewidth}{0.5pt}
    \stackinset{l}{57pt}{t}{0in}%
    {\rotatebox{90}{\hdashrule{62pt}{0.5pt}{0.3mm}}}{%
        \includegraphics[width=0.99\textwidth]{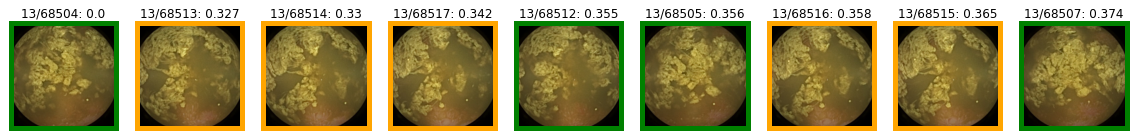}
    }%
    \hspace{-\linewidth}\rule{\linewidth}{0.5pt}
    \stackinset{l}{57pt}{t}{0in}%
    {\rotatebox{90}{\hdashrule{62pt}{0.5pt}{0.3mm}}}{%
        \includegraphics[width=0.99\textwidth]{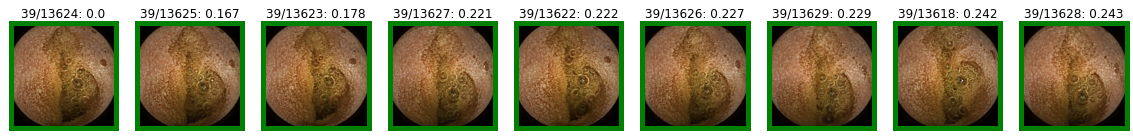}
    }
    
    \caption{Given samples from the test set, shown in the first column, each row represents other samples in the set sampled by distance in the embedding space. Each image is titled as \textit{video/frame: distance}, and framed in red if they come from a different video, orange if it is the same video, and green if, additionally to being in the same video, they are within $w$ distance.}
    
    \label{fig:embeddings}
\end{figure*}

Taking into account we measure similarity with euclidean distance in the TL function, two embeddings are considered close if their distance is relatively near the margin parameter, or distant otherwise. As can be seen from Figure \ref{fig:embeddings}, the network successfully distinguishes not only images that are completely different but also correctly represents images that are similar while not being consecutive.

Similarly, some samples are close to frames of other videos while maintaining evident similarities, which serves to justify that the network has not learned features specific to a video, but, rather, it has trained for rich information. Our time-based contrastive learning implicitly enables the model to identify similarities between different videos with similar events, which is vital for SSL, as the finetune process needs this augmented information to properly function.

\begin{figure}
    \centering
    \includegraphics[width=\linewidth]{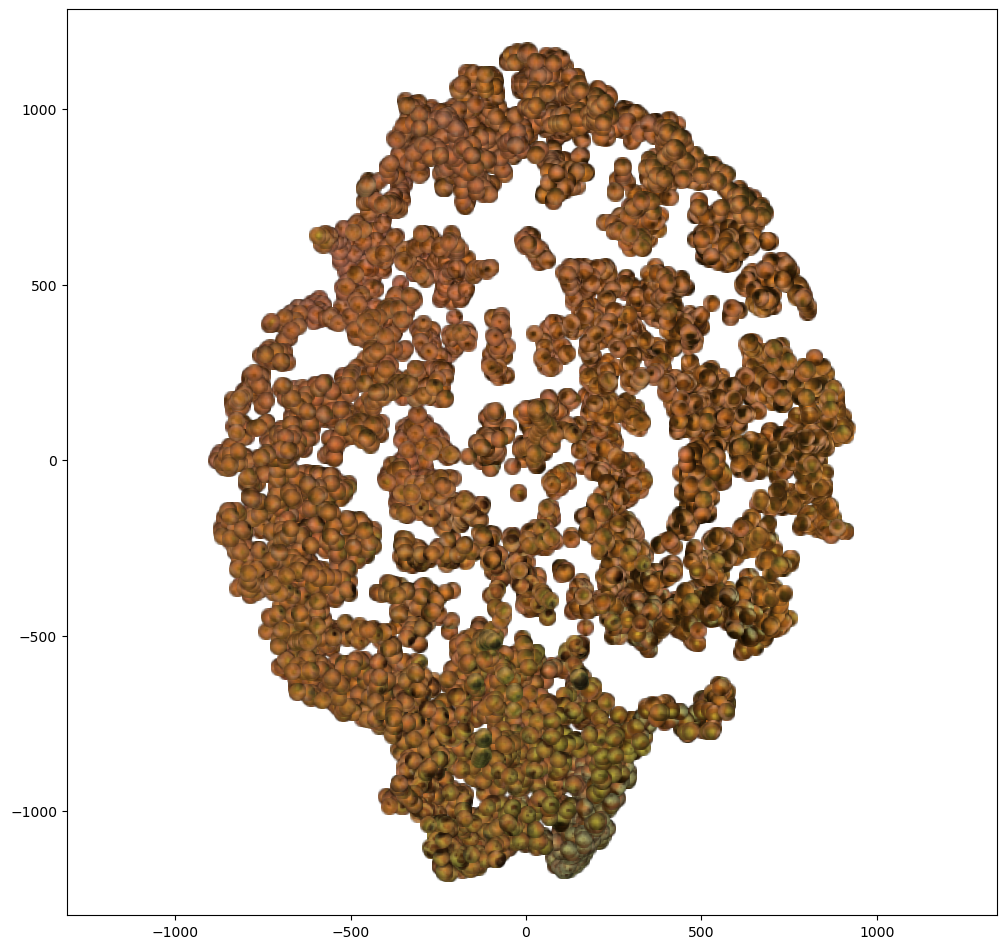}
    \caption{t-SNE of the embeddings post-projections obtained from one WCE video after the pretrain phase. The representation shows that visually alike images have close embeddings.}
    \label{fig:tsne}
\end{figure}

To further validate the embeddings, we obtained a t-SNE representation \cite{VanDerMaaten2008} of one WCE video. As can be seen in Figure \ref{fig:tsne}, frames that are visually close, containing for example similar structures and colors, are densely packed in the same area of the representation. This indicates that their embeddings are also close, verifying that the network is learning our contrastive metric successfully.

\subsubsection*{Polyp dataset}
Following previous work from Laiz \textit{et al.} \cite{Laiz2020a}, we abandon traditional metrics used in polyp detection. Accuracy, for instance, is a skewed metric under such data imbalances, favoring the class with most examples in detriment to the overall performance. Thus, as proposed in their publication, we adopt AUC ROC as the primary metric.

Moreover, sensitivity at set specificity thresholds, namely 95\%, 90\%, and 80\%, are also reported. Not only are they robust towards imbalance, but most importantly they provide helpful information regarding the number of images a physician needs to check to obtain a certain level of performance in polyp detection. For instance, this metric gives a measure of how many polyps would be detected if a percentage of negatives was discarded based on the classifier.

To ensure that similar images, which are commonly found in sequential frames in videos, are not present in both train and evaluation simultaneously, we split the dataset based on whole videos. Consequently, a patient can only be found either in train or evaluation, but never in both. Failing to do so would overestimate the performance, producing better results while probably failing to generalize with new data.

The baseline for this particular task, further referred to as Imagenet, uses a ResNet-50 preinitialized with Imagenet and trained on this same dataset. Unlike our model, the Imagenet model uses no SSL nor any contrastive loss. A more advanced model, $TL_{BA}$ as trained in Laiz \textit{et al.} \cite{Laiz2020a}, introduces a TL to the previous model. Finally, the state-of-the-art contrastive learning architecture SimCLR \cite{Chen2020b}, is also compared.

Every result, as seen in Table \ref{table:results}, is reported as the mean value and standard deviation obtained from a 5-fold cross-validation. Each evaluation set is done with whole videos, not individual samples. Also, each fold is finetuned and evaluated independently, starting from exactly the same initial values taken from our pretrained network.

Adding any kind of contrastive losses, as can be seen from $TL_{BA}$ and SimCLR, already provides a significant boost of 10\% on the AUC score over the baseline. However, our method based on SSL outperforms the former models by close to a $2\%$. This significant improvement can be observed across all metrics, meaning SSL and our particular time-based contrastive learning can extract information that remains otherwise hidden or ignored. Of particular interest are the improvements in the sensitivity at different specificities. Our method can give a notable increase in the number of polyps correctly classified when discarding varying amounts of negatives. 

\begin{figure}
    \flushright

    \begin{subfigure}{.225\textwidth}
        \makebox[0pt][r]{\makebox[30pt]{\raisebox{40pt}{(a)}}}%
        \centering
        \includegraphics[width=\linewidth]{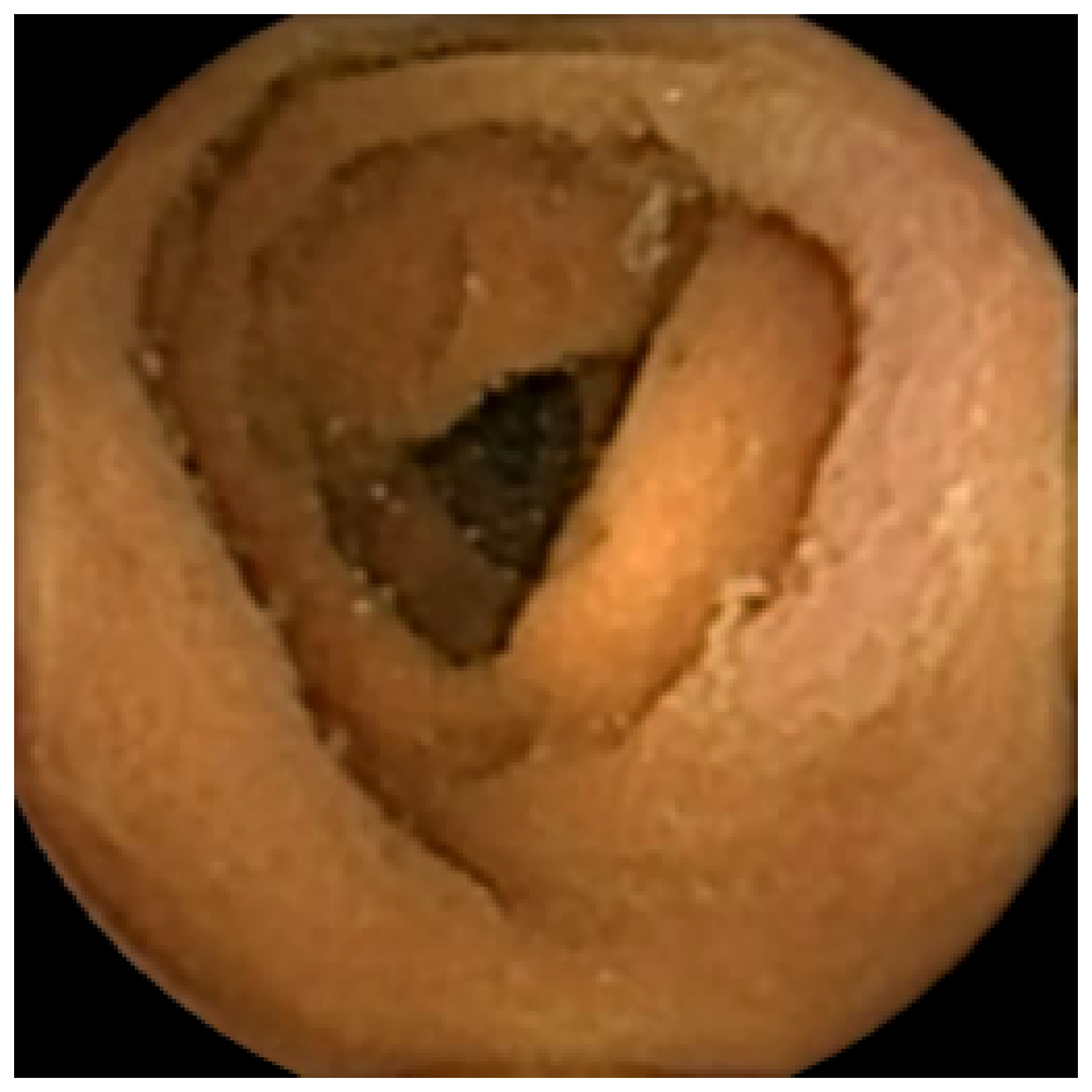}
        \label{fig:false-pos-1}
    \end{subfigure}%
    \begin{subfigure}{.225\textwidth}
        \centering
        \includegraphics[width=\linewidth]{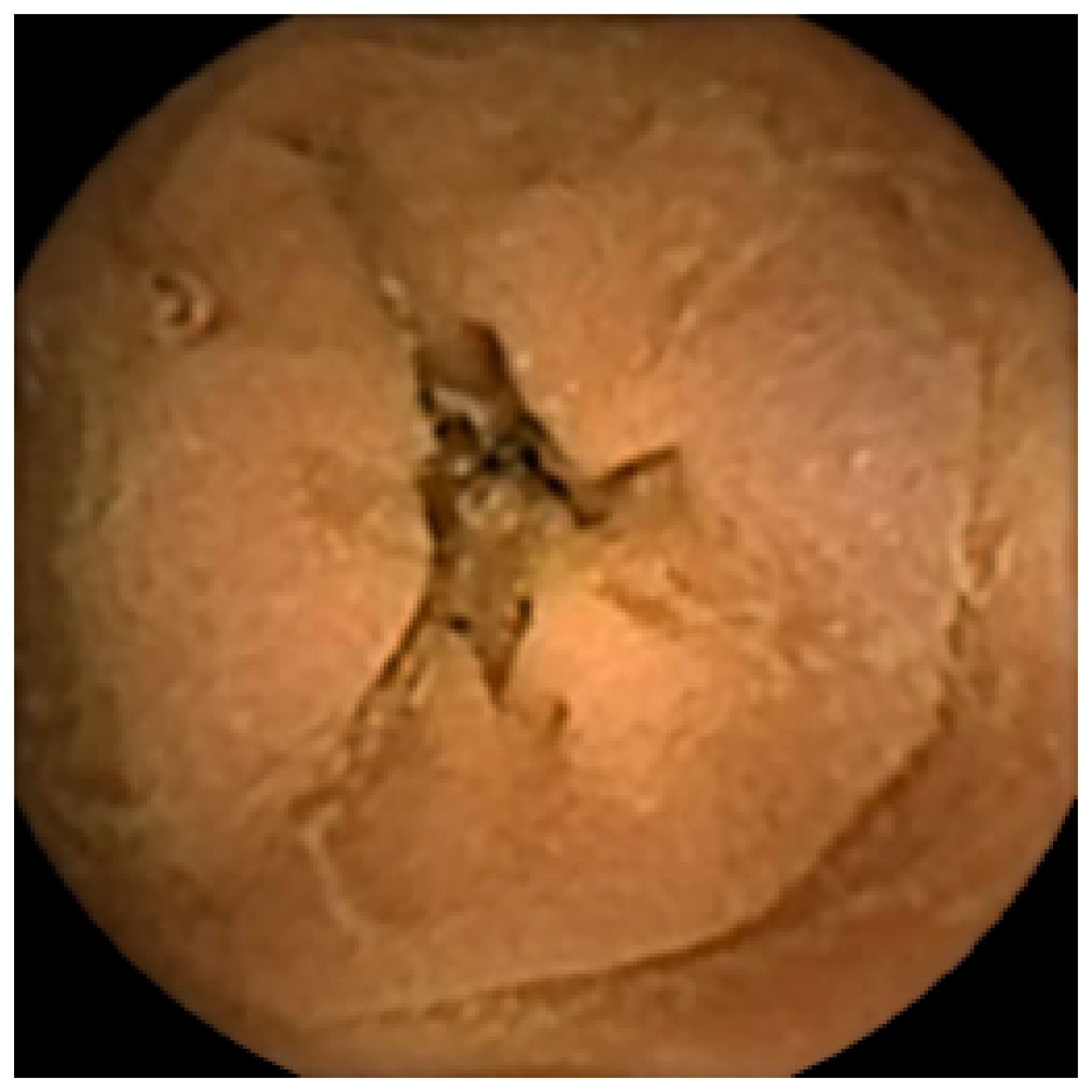}
        \label{fig:false-pos-2}
    \end{subfigure}

    \begin{subfigure}{.225\textwidth}
        \makebox[0pt][r]{\makebox[30pt]{\raisebox{40pt}{(b)}}}%
        \centering
        \includegraphics[width=.99\linewidth]{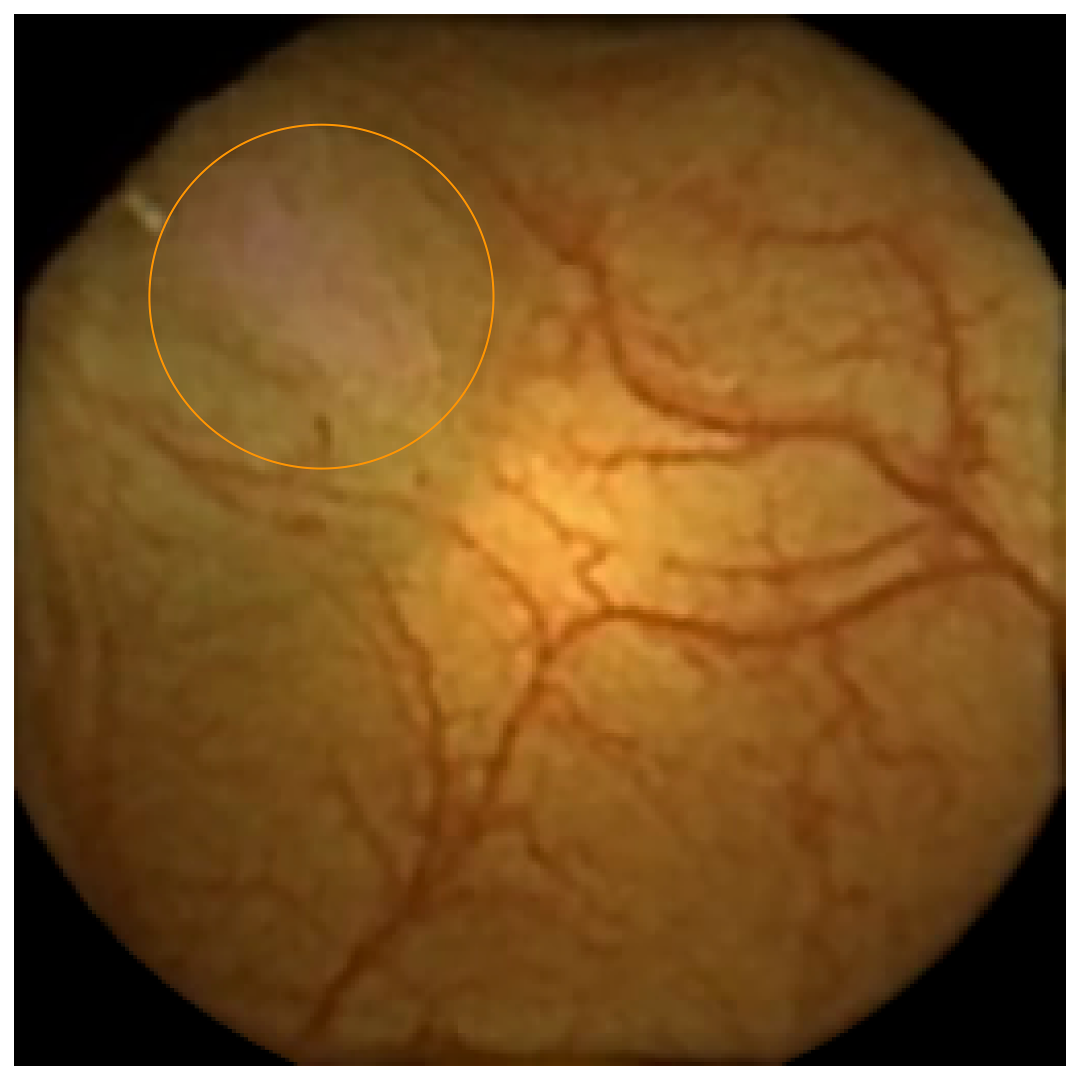}
        \label{fig:false-neg-1}
    \end{subfigure}%
    \begin{subfigure}{.225\textwidth}
        \centering
        \includegraphics[width=.99\linewidth]{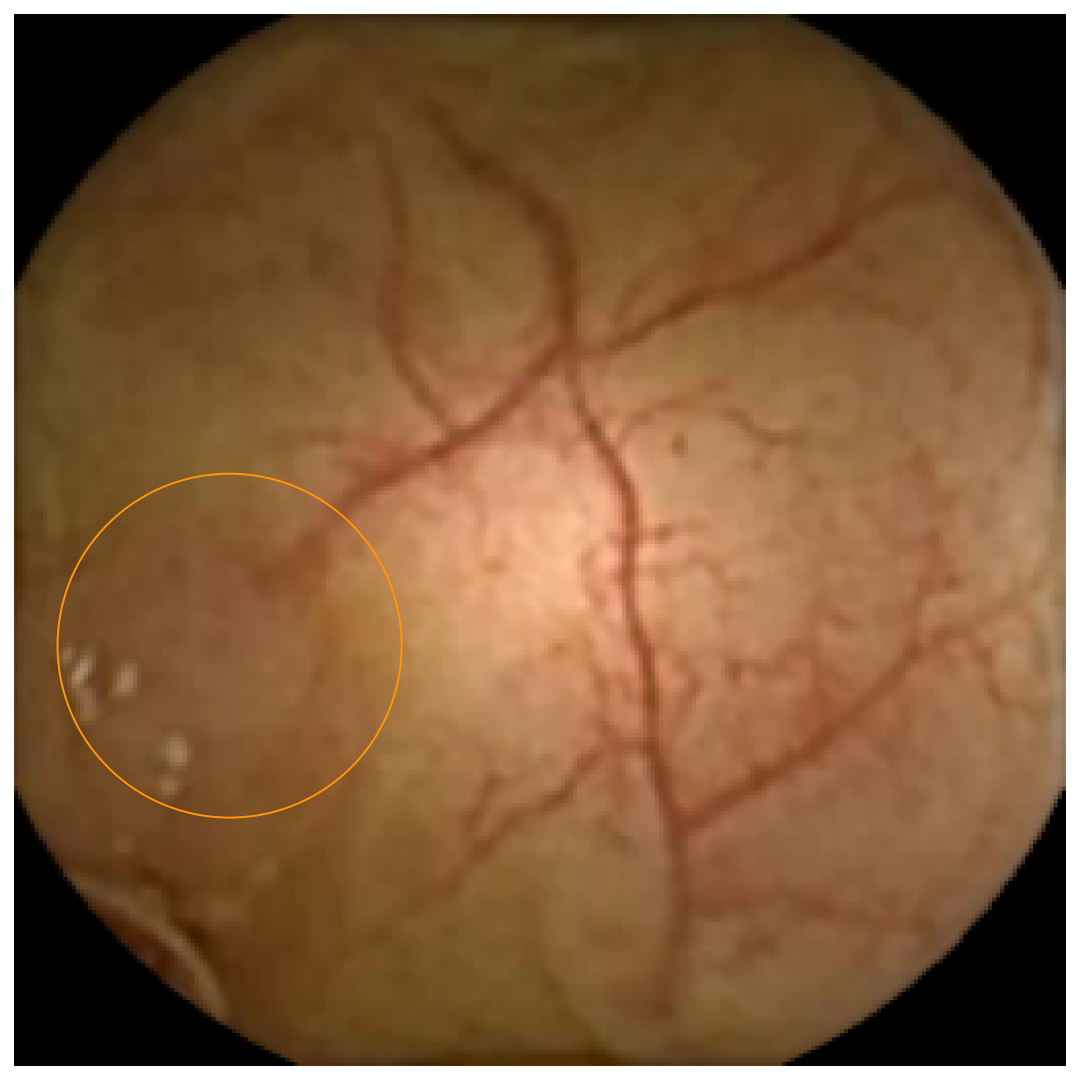}
        \label{fig:false-neg-2}
    \end{subfigure}
    
    \caption{Random samples from the test set. Row a) shows two false positives, images inaccurately classified as polyps. Row b) depicts two false negatives. The polyps have been circled to help with their identification.}
    \label{fig:posneg}
\end{figure}

Another approach to validation, aside from the quantitative analysis above, is to inspect and visualize the results. In other words, performing a qualitative validation of the results by examining where the model is performing correctly and where it is failing. Miss-classified non-polyp images would add more work to the physician due to having to unnecessarily check false positives. However, not showing a polyp frame because the system has falsely classified it as negative can have a devastating effect, with implications much severe than its counterpart case. Figure \ref{fig:posneg} depicts two examples of the mentioned cases. It can be seen that the network fails in especially tough cases, where the polyp would be hard to be seen even for a physician. The polyps have been circled for the reader to identify where they are. False positives occur in zones with a more pinkish tone, characteristic of polyps, and always in rugged and wrinkled surfaces, which could explain why the network is mistaking them for polyps.

\subsubsection*{CAD-CAP WCE}

Following the procedure established in \cite{Guo2020a}, we have split the data into 4 sets and performed a 4-fold cross-validation. As per the original challenge \cite{Dray2018}, we report in Table \ref{table:giana-results} the per-class Matthews correlation coefficient (MCC) and F1 scores, and the overall accuracy as $p_0$.

A naive implementation, using a ResNet-50 and without SSL, fails to correctly classify a significant portion of the data, achieving only a $69.98\%$ accuracy. However, adding SSL to this same model and using the method we propose in this publication, immediately boosts every metric by more than $20\%$. Our implementation reaches a total of $92.77\%$ accuracy without any change to the architecture.

Further, we compare our results with those reported by Guo \textit{et al.} \cite{Guo2020a}, the current state-of-the-art model for CAD-CAP. They handcrafted a network for this dataset and provide six baselines and one additional model that uses semi-supervision to improve the results. With respect to the baselines, our model obtains higher scores across most metrics, as can be observed in Table \ref{table:giana-results}. We also attain comparable results to their best implementation, which has a semi-supervised phase training over 1807 unlabeled images provided by CAD-CAP that we do not use.

\begin{table*}[t]

    \caption{Per class and overall results of various methods in GIANA. ResNet is the same architecture as Ours but without the SSL step. Baseline 1 and 6 refer to the baselines reported by Guo \textit{et al.} \cite{Guo2020a}, while the model with the same name is their semi-supervised performing implementation.}
    
    \centering
    
    \begin{tabular}{ccccc}
    \hline
    \textbf{Method}           & \textbf{Class} & \textbf{F1-Score (\%)}  & \textbf{MCC (\%)}       & \textbf{$\mathbf{p_0}$ (\%)}             \\ \hline
    \multirow{3}{*}{ResNet}   & Normal         & $73.28\pm3.57$          & $60.58\pm5.44$          & \multirow{3}{*}{$69.98\pm1.35$}          \\
                              & Inflammatory   & $65.19\pm2.95$          & $55.86\pm1.77$          &                                          \\
                              & Vascular       & $70.79\pm4.60$          & $65.35\pm3.80$          &                                          \\ \hline
    \multirow{3}{*}{Baseline 1 \cite{Guo2020a}} & Normal         & $94.92\pm0.71$          & $92.37\pm1.07$          & \multirow{3}{*}{$84.99\pm0.80$}          \\
                              & Inflammatory   & $79.24\pm1.55$          & $68.72\pm2.15$          &                                          \\
                              & Vascular       & $80.75\pm1.65$          & $71.49\pm2.57$          &                                          \\ \hline
    \multirow{3}{*}{Baseline 6 \cite{Guo2020a}} & Normal         & $96.41\pm0.84$          & $94.61\pm1.26$          & \multirow{3}{*}{$91.92\pm1.71$}          \\
                              & Inflammatory   & $88.98\pm2.13$          & $83.44\pm3.24$          &                                          \\
                              & Vascular       & $90.27\pm2.78$          & $85.75\pm3.73$          &                                          \\ \hline
    \multirow{3}{*}{Ours}     & Normal         & $95.00\pm1.13$ & $92.57\pm1.66$ & \multirow{3}{*}{$92.77\pm1.20$} \\
                              & Inflammatory   & $89.87\pm1.65$ & $84.99\pm2.46$ &                                          \\
                              & Vascular       & $90.26\pm1.76$ & $85.78\pm2.37$ &                                          \\ \hline
    \multirow{3}{*}{Guo \textit{et al.} \cite{Guo2020a}}     & Normal         & $97.41\pm0.45$ & $96.10\pm0.69$ & \multirow{3}{*}{$93.17\pm1.14$} \\
                              & Inflammatory   & $90.30\pm1.56$ & $85.43\pm2.24$ &                                          \\
                              & Vascular       & $91.69\pm1.21$ & $87.78\pm2.06$ &                                            
    \end{tabular}
    
    \label{table:giana-results}

\end{table*}

%
%

%
%

\section{Conclusion}

In this work, we propose an SSL method that leverages the information in the temporal axis of WCE videos to obtain rich embeddings. Our method introduces a pseudo-labeling process that enables time-based contrastive learning, images close in a video be represented by similar embeddings.

We demonstrate that using this process yields better results in subsequent models specializing in domain-specific tasks. For instance, we test the method to detect several events in the GIANA dataset, obtaining comparable results to state-of-the-art models with reduced complexity and parameter count at a 92.77\% accuracy. Similarly, using the SSL model to classify polyps shows an increase in successful polyp detection, reaching a 95.00\% AUC.

Thus, we claim that using SSL when leveraging temporal information is beneficial for WCE models. Most importantly, the method imposes no requirements for the dataset used during the supervised phase, effectively tackling the classical problems commonly encountered in medical imaging: low amounts of data—specially labeled—and severe class imbalances.

Overall, we strongly believe the method is a good step towards better models that empower CADx models in medical interventions. For instance, a higher rate of polyp detection would decrease the time spent by physicians revising WCE videos, allowing more accurate diagnosis in shorter amounts of time.

Future work could focus on exploring other SSL architectures that might boost the downstream tasks' performance. Moreover, expanding the method to other WCE domains and other medical fields would also be of high interest.

\section*{Acknowledgments}
This work was partially founded by MINECO Grant RTI2018-095232-B-C21, SGR 1742, Innovate UK project 104633, and by an FPU grant (Formacion de Profesorado Universitario) from the Spanish Ministry of Universities to Guillem Pascual (FPU16/06843). We gratefully acknowledge the support of NVIDIA Corporation with the donation of the Titan Xp Pascal GPU used for this research.

\bibliography{Self-Supervised-Paper}

\end{document}